\documentclass[10pt,journal,compsoc]{IEEEtran}
\ifCLASSOPTIONcompsoc
  \usepackage[nocompress]{cite}
\else
  \usepackage{cite}
\fi

\usepackage{graphicx}
\usepackage[colorlinks=true,linkcolor=blue,citecolor=blue]{hyperref}
\usepackage{multirow}
\usepackage{amsmath}
\usepackage{amssymb}
\usepackage{url}
\usepackage{arydshln}

\ifCLASSINFOpdf
\else
\fi
\begin{document}
%
\title{Training Convolutional Neural Networks and Compressed Sensing End-to-End for Microscopy Cell Detection}
%
%
%
%

\author{Yao~Xue,
        Gilbert~Bigras,
        Judith~Hugh,
        Nilanjan~Ray
\IEEEcompsocitemizethanks{\IEEEcompsocthanksitem Yao~Xue and Nilanjan~Ray are with the Department of Computing Science, University of Alberta. E-mail: yxue2@ualberta.ca, nray1@ualberta.ca
	
\IEEEcompsocthanksitem Judith~Hugh and Gilbert~Bigras are with the Cross Cancer Institute, University of Alberta. E-mail: Judith.Hugh@albertahealthservices.ca, Gilbert.Bigras@albertahealthservices.ca.\protect\\}


}

%
%

\markboth{IEEE Transactions on Medical Imaging, 2018}%
{Shell \MakeLowercase{\textit{et al.}}: Bare Demo of IEEEtran.cls for Computer Society Journals}
%



\IEEEtitleabstractindextext{%
\begin{abstract}

Automated cell detection and localization from microscopy images are significant tasks in biomedical research and clinical practice. In this paper, we design a new cell detection and localization algorithm that combines deep convolutional neural network (CNN) and compressed sensing (CS) or sparse coding (SC) for end-to-end training. We also derive, for the first time, a backpropagation rule, which is applicable to train any algorithm that implements a sparse code recovery layer. The key observation behind our algorithm is that cell detection task is a point object detection task in computer vision, where the cell centers (i.e., point objects) occupy only a tiny fraction of the total number of pixels in an image. Thus, we can apply compressed sensing (or, equivalently sparse coding) to compactly represent a variable number of cells in a projected space. Then, CNN regresses this compressed vector from the input microscopy image. Thanks to the SC/CS recovery algorithm ($L_1$ optimization) that can recover sparse cell locations from the output of CNN. We train this entire processing pipeline end-to-end and demonstrate that end-to-end training provides accuracy improvements over a training paradigm that treats CNN and CS-recovery layers separately. Our algorithm design also takes into account a form of ensemble average of trained models naturally to further boost accuracy of cell detection. We have validated our algorithm on benchmark datasets and achieved excellent performances.

\end{abstract}

\begin{IEEEkeywords}
Cell Detection, Convolutional Neural Network, Backpropagation, Compressed Sensing, sparse coding, end-to-end training.
\end{IEEEkeywords}}

\maketitle

\IEEEdisplaynontitleabstractindextext

%
\IEEEpeerreviewmaketitle

\IEEEraisesectionheading{\section{Introduction}
	\label{sec:introduction}}

%
%
%
%

\begin{figure}[htbp]
	\centering
	\setlength{\abovecaptionskip}{0pt}
	\setlength{\belowcaptionskip}{0pt}
	\includegraphics[width=5cm]{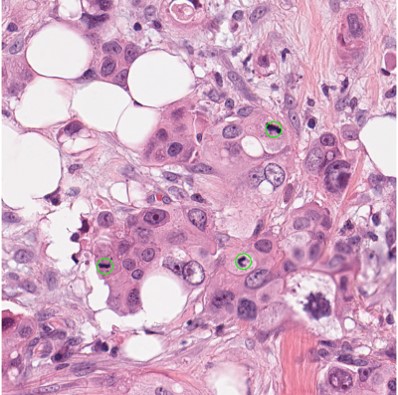}
	\caption{A microscopy image with three annotated target cells.}
	\label{fig:cell_intro}
\end{figure}

\IEEEPARstart{D}{etection} and localization of certain types of cells or nuclei from microscopy images are of significant importance to clinical practices, medical and biomedical research, as well as computer vision. For example, in the context of breast cancer, the percentage of proliferating (e.g. Ki-67 positive) tumor cells is utilized to establish the prognosis of the disease \cite{YaoICIP}. In order to access the percentage, pathologists eyeball cells of interest in histological slides under a microscope that is quite painstaking and prone to fatigue. Fig. \ref{fig:cell_intro} shows a microscopy image with human annotated target cells. A computer program is expected to replace the human annotation task. Automated cell detection and localization act as a benchmark task for computer vision research, as evidenced by several competitions organized by medical imaging societies \cite{AMIDA-2013,AMIDA-2016,ICPR-2012,ICPR-2014}.

Automated cell detection and localization have several challenges. First, the target cells are often surrounded by lookalike clutters, which forces an algorithm to generate false positives. Second, the target cells may vary significantly in appearances. Third, target cells may appear very sparsely (such as in tens), moderately densely (in hundreds) or very densely (in thousands) in a high resolution microscopy image \cite{YaoICIP}. Fourth, sometimes there is significant intensity variations in the background as well as in the stains present in microscopy images.

\begin{figure}[htbp]
	\centering
	\setlength{\abovecaptionskip}{0pt}
	\setlength{\belowcaptionskip}{0pt}
	\includegraphics[width=8.5cm]{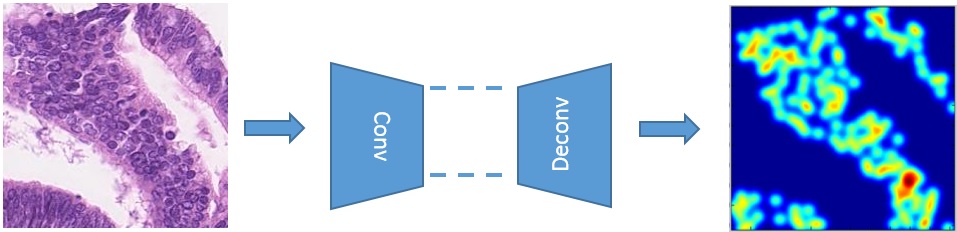}
    (a)
	\includegraphics[width=8.5cm]{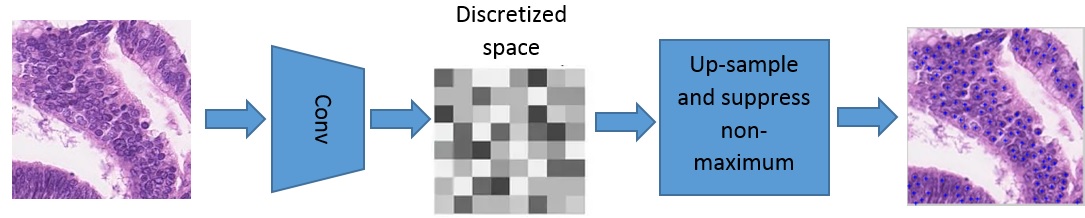}
	(b)
	\includegraphics[width=8.5cm]{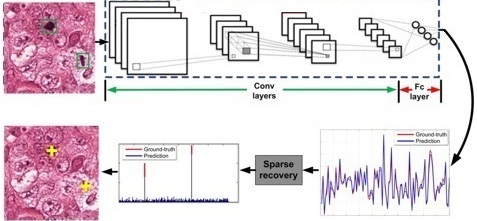}
	(c)
	\caption{(a) An example density prediction architecture \cite{Zisserman2018}; (b) An example space discretization architecture \cite{CasNN}; (c) Proposed end-to-end CNNCS architecture.}
	\label{fig:architectures}
\end{figure}

Even though pattern recognition has advanced significantly in the recent past, the aforementioned challenges pose significant roadblocks towards designing commercial software for automated cell detections. Due to the advancement of deep learning in visual recognition, state-of-the-art methods for cell detection and localization are all based on convolutional neural networks, as evidenced by cell detection challenges \cite{AMIDA-2016,ICPR-2014}. In general, these methods fall into two categories as shown in Fig. \ref{fig:architectures}(a) and (b). The first category predicts a \textbf{density function} on the pixel space, where the peaks of the density correspond to the cell centers. An example of the density prediction method has been published in 2018 \cite{Zisserman2018}. In the second category, an \textbf{explicit space discretization} mechanism is applied on the output pixel space to detect cell locations. The space discretization may be wrapped underneath state-of-the-art object detection techniques, such as faster rcnn~\cite{faster-rcnn} or YOLO~\cite{yolo2016}. An example method that uses space discretization has been published in 2016 \cite{CasNN}.

We argue that the existing methods have some limitations. Density prediction methods, during their training, measure pixel-by-pixel deviation of the output density from a ground truth density in their training process, and become excessively sensitive to the smoothing method used in creating the ground truth density from the point (cell center) annotations. If cells are densely present, peaks in the density function tend to merge underestimating the cell locations. Furthermore, sparse cell locations create imbalance in the cost function between positive and negative pixel locations.

Explicit space discretization partitions the output pixel space and hence suffers from a loss of resolution, especially, when cells are densely present in an image. Explicit space discretization-based methods are better in dealing with sparsely present cells, because originally these methods were created to detect and localize extended objects, such as cars, people, etc. that appear sparsely in a natural scene. However, explicit space discretization still suffers from the class imbalance problem resulting in false positives, and as a result they need to be combined with a cascaded layer of classifier to filter out false positives \cite{CasNN}.

Thus, we notice that existing methods for automated cell detection suffer from two types of issues: discretization of output space and the class imbalance issue. Based on our observations, we argue that a \textbf{fundamentally new algorithm} design is required to deal with the point object detection, such as microscopy cell detection and localization. In this work, the novel theme we use for cell (point object) detection and localization is \textbf{compression instead of discretization}. A schematic in Fig. \ref{fig:architectures}(c) explains our approach. We refer to our method as CNNCS (convulutional neural net + compresses sensing). In CNNCS, the CNN predicts a fixed length vector from an input microscopy image. Then, a sparse reconstruction layer (compressed sensing (CS)/sparse coding(SC)) recovers the sparse cell locations in the pixel space. We illustrate that CNNCS can be trained end-to-end. Due to CS, sparse cell locations are projected to a fixed length vector and there is no space discretization in CNNCS, unlike all previous methods. Also, the loss function in CNNCS is a combination of $L_2$ and $L_1$ loss for regression, as opposed to classification loss and hence alleviates the class imbalance issues to some extent. In addition, we show that CNNCS can have an implicit ensemble averaging of models to boost the accuracy of cell detection.

From the perspective of computing science research, this is the first work that derives a backpropagation rule for a CS/SC layer. This backpropagation is independent of the algorithm used for the CS/SC recovery. Thus, our derivation opens up new opportunities to utilize differentiable and learn-able SC/CS within a larger end-to-end learning program. The only known previous differentiable SC used recurrent neural networks to approximate two specific iterative $L_1$ recovery algorithms \cite{LISTA}, unlike a general version of the backpropagation we derive in this paper. 

Our previous works \cite{YaoICIP,AAAI_workshop} have introduced the concept of combining CNN with CS for cell detection. The presented framework here extends our previous work by introducing end-to-end training and a derivation of backpropagation across compressed sensing layer. In summary, our contributions are as follows:
\begin{itemize}
\item This is the first attempt to combine CNN with CS in an \textbf{end-to-end}, or, equivalently, in a differentiable fashion to solve the cell detection and localization.
\item By using CS in our algorithmic framework, we do not discretize the pixel space, instead we compress it.
\item CNNCS mitigates the class imbalance issue by converting a classification task into a regression task.
\item CNNCS is able to introduce a form of implicit model ensemble to boost accuracy.
\item CNNCS achieves excellent results on the benchmark cell detection datasets when compared with the state-of-the-art methods.
\item Lastly, we derive a general backpropagation rule across a compressed sensing/sparse coding layer that is independent of the algorithm used in the layer.
\end{itemize}



\section{Background and Related Work} \label{bg}
\subsection{Cell Detection and Localization}
In the last few decades, different cell recognition methods had been proposed \cite{Meijering12}. Traditional computer vision based cell detection systems adopt classical image processing techniques, such as intensity thresholding, feature detection, morphological filtering, region accumulation, and deformable model fitting. For example, Laplacian-of-Gaussian (LoG) \cite{LoG} operator was a popular choice for blob detection; Gabor filter or LBP \cite{LBP} offers many interesting texture properties and had been attempted for cell detection \cite{LoG-cellSeg}.

Conventional cell detection approaches follow a ``hand-crafted feature representation''+``classifier'' framework. First, detection system extracts one (or multiple) kind of features as the representation of input images. Image processing techniques offer a range of feature extraction algorithms for selection. After that, machine learning based classifiers work on the feature vectors to identify or recognize regions containing target cells. ``Hand-crafted feature representation''+``classifier'' suffers from some limitations. It is a non-trivial task for humans to select suitable features. In many cases, it requires significant prior knowledge about the target cells and background. What is more important, the performance of a hand-crafted feature-based classifier soon reaches an accuracy plateau, even when trained with plenty of training data.

In comparison to the conventional cell detection methods, deep neural networks recently has been applied to a variety of computer vision problems, and has achieved better performance on benchmark vision datasets \cite{AMIDA-2016}. The most compelling advantage of deep learning is that it has evolved from fixed feature design strategies towards automated learning of problem-specific features directly from training data \cite{LeCun-nature}. By providing massive amount of training images and problem-specific labels, users do not have to go into the elaborate procedure for the extraction of features.

The state-of-the-art methods in detection and localization today include deep learning techniques. Soon after Fully Convolutional Network (FCN) \cite{FCN} was proposed for semantic segmentation, Xie et al. \cite{Alpher25} presented a FCN-based framework for cell counting, where their FCN is responsible for predicting a spatial density map of target cells, and the number of cells can be estimated by an integration over the learned density map. A convolutional autoencoder architecture \cite{Zisserman2018} has been introduced for cell density map prediction. Chen et al. \cite{CasNN} uses an object detector for candidate region selection, and then another CNN for further discrimination between target cells and background.

Cirean et al. \cite{DNN-IDSIA} proposed a mitosis detection method by CNN-based prediction followed by ad-hoc post processing. As the winner of ICPR 2012 mitosis detection competition, they used deep max-pooling convolutional neural networks to detect mitosis in breast histology images \cite{DNN-IDSIA}. The networks were trained to classify each pixel using a patch centered on the pixel as a context. Then post processing was applied to the network output. Shadi et al. \cite{AggNet16} used expectation maximization within a deep learning framework in an end-to-end fashion for mitosis detection. This work presents a new concept for learning from crowd sourcing that handles data aggregation directly as part of the learning process of the convolutional neural network (CNN) via additional crowd-sourcing layer. It is the first piece of work where deep learning has been applied to generate a ground-truth labeling from non-expert crowd annotation in a biomedical context.

\subsection{Compressed Sensing-based Output Encoding}

Compressed sensing or compressive sensing (CS) \cite{Alpher17}, \cite{Alpher18}, \cite{Alpher16} and sparse coding (SC) \cite{Elad2010} have emerged as new frameworks for signal acquisition and reconstruction, with rich theoretical results and significant practical applications, such as MRI scan time reduction \cite{BirnsKKSN16} and cheaper camera design \cite{Duarte2008}. 

In CS/SC, a \textbf{sparse signal} $a$ is sensed by a limited number of linear observations $x$:
\begin{equation} \label{eq:cs}
x= D a,
\end{equation}
where $D$ is a $m \times n$ sensing matrix, with typically $m \ll n.$ CS theory \cite{Alpher17,Alpher18} states that given $x$ and $D$, a convex optimization can recover $a,$ provided the sensing matrix $D$ satisfies a restricted isometry property (RIP) and $m \geq  C_m k log(n)$, where $C_m$ is a small constant greater than one and $k$ is the maximum number of non-zero elements in $a.$ 

Given $D$ and $x$, the recovery of $a$ typically relies on a convex optimization with a penalty expressed by $L_1$ norm as follows:
\begin{equation} \label{eq:sp1}
\underset{a}{\text{min}} \ \frac{1}{2}\| Da-x \|_2^{2}+\lambda\| a \|_1, 
\end{equation}
where $\lambda$ is a non-negative weight balancing the two terms in the cost function (\ref{eq:sp1}). Various algorithms exist today that can optimize (\ref{eq:sp1}). Examples include orthogonal matching pursuit (OMP) \cite{OMP} and dual augmented Lagrangian (DAL) method \cite{DAL}. 

Cell centroids are point objects sparsely dispersed in an image. Thus, we apply CS-based encoding of the cell locations that compresses sparse cell locations $a$ into a much smaller dense vector $x$ with a sensing matrix $D.$ Such transformations are known as output encoding \cite{AAAI_workshop} that can sometimes generate more accurate predictions. Successful examples of output encoding include error corrections \cite{ECOC} and redundancy in the output representation \cite{RAkEL}.

The steps for CS-based output encoding is as follows. First, a sparse label $a$ is projected to a short and dense vector $x$. Then, a machine learner regresses $x.$ Finally, a recovery algorithm, such as OMP or DAL is applied to recover $a$ using minimization (\ref{eq:sp1}). 

CS-based output encoding has a rather modest presence in the literature, where it was applied with linear and non-linear machine learners. It started with the work of Hsu et al. \cite{CS} that has proved a generalization bound for such methods. The generalization prediction error is bound by two factors: how well the machine learner has predicted and how well the recovery algorithm has worked.

More recently, non-linear predictors such as Bayesian learner \cite{Bayesian-CS}, decision trees \cite{output-space-thesis} and CNN \cite{YaoICIP,AAAI_workshop} were used. Our work presented in this paper is more comprehensive and it expands on our previous endeavors \cite{YaoICIP,AAAI_workshop}. Here we present a novel \textbf{end-to-end} framework for cell (point object) detection with CS-based output encoding. The end-to-end framework has become ubiquitous in the deep learning community; so, it is a natural extension in that sense. Also, given the generalization bound \cite{CS}, it makes more sense to optimize both the prediction and recovery simultaneously in an end-to-end fashion.

\section{Proposed Method}

\subsection{Cell Location Encoding} \label{sec:encoding-scheme}

\begin{figure}[h]
	\centering
	\setlength{\abovecaptionskip}{0pt}
	\setlength{\belowcaptionskip}{0pt}
	\includegraphics[width=9cm]{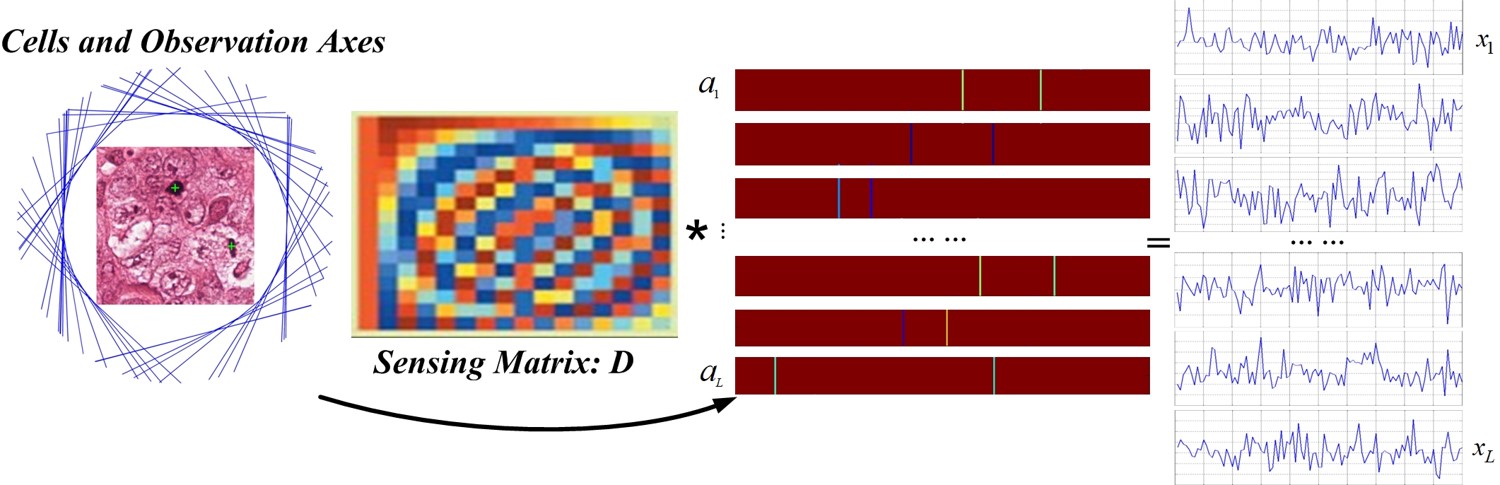}
	\caption{Cell location encoding by random projections \cite{AAAI_workshop}.}
	\label{encoding1}
\end{figure}

CNNCS relies on encoding cell center locations into a dense code, which a CNN predicts from an input image. We use a form of encoding, which we refer to as encoding by random projections \cite{AAAI_workshop}, as shown in Fig.~\ref{encoding1}. A number of (say, $L$) straight lines are created around an image. The orientations of these straight lines are uniformly distributed. All cell centroid locations (pixel coordinates) are then projected onto these lines, so that signed distances from the straight line denote the cell locations. Note that this type of encoding creates \textbf{redundancy} in the cell location representation, as there are multiple lines encoding the cell locations. In the rare event, two or more cells may be projected to the same location on the a straight line. However, because of the redundant representations, cell centrods can be still recovered using other signed distances.

To describe encoding by random projections, let $a_l$ denote the signed distances of cells on the $l^{th}$ straight line. If the image patch size is $h \times w,$ then the length of vector $a_l$ is $n=\sqrt[]{h^2+w^2}.$ We create dense codes of length $m$: $x_l = Da_l.$ Concatenating $x_l, l=1,2,...,L$, into a single vector $x$ completes this encoding process.

\subsection{End-to-End CNNCS}

\begin{figure}[h]
	\centering
	\setlength{\abovecaptionskip}{0pt}
	\setlength{\belowcaptionskip}{0pt}
	\includegraphics[width=8.5cm]{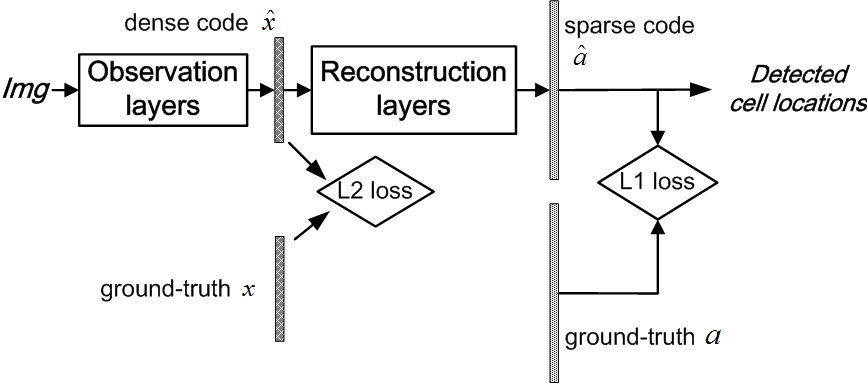}
	\caption{End-to-end training framework for CNNCS.}
	\label{end-to-end}
\end{figure}

\begin{figure}[h]
	\centering
	\setlength{\abovecaptionskip}{0pt}
	\setlength{\belowcaptionskip}{0pt}
	\includegraphics[width=8.5cm]{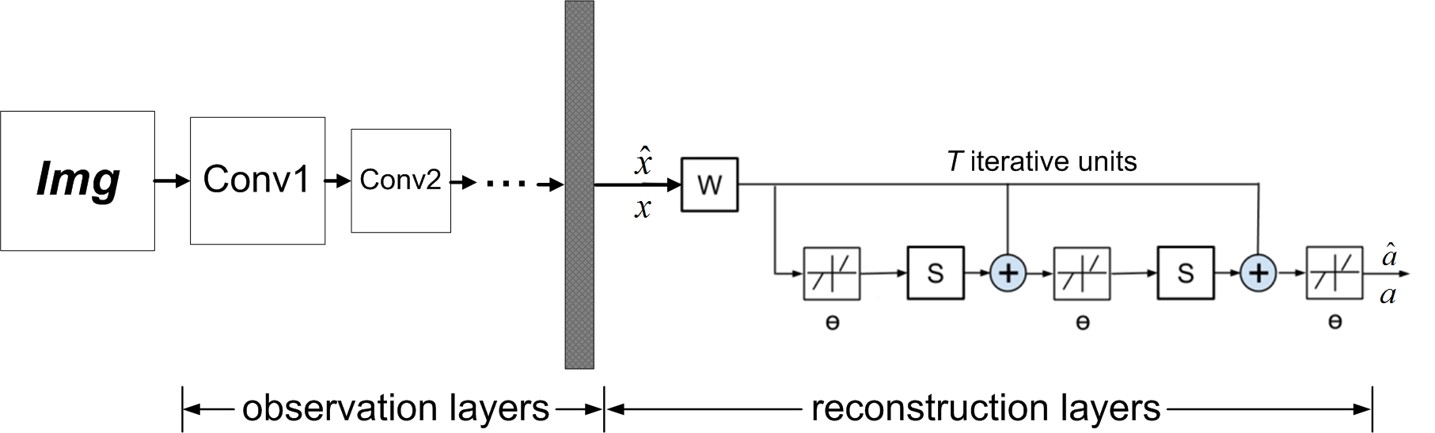}
	\caption{End-to-end architecture for CNNCS: CNN + LISTA.}
	\label{end-to-end-net}
\end{figure}

Fig.~\ref{end-to-end} shows End-to-End CNNCS framework. The input image goes through an observation layer composed of a CNN that outputs a dense vector $\hat{x},$ which is compared to the encoded, ground-truth dense vector $x=Da$ by $L_2$ norm. Note that $D$ is the $m\times n$ random Gaussian projection matrix (each element is i.i.d. zero mean Gaussian with variance $1/m$) and $a$ is the ground-truth sparse cell location vector $a.$ Note that the previous subsection describes encoding of $a.$ The predicted dense vector $\hat{x}$ is then fed to a recovery/reconstruction layer that reconstructs a sparse output vector $\hat{a},$ which is compared to $a$ by $L_1$ norm. Thus, the cost function for End-to-End CNNCS is a mixture of $L_2$ and $L_1$ norms as follows:
\begin{equation} \label{eq:overall_loss}
loss = \frac{1}{2}\| \hat{x}-x \|_2^{2}+\alpha\| \hat{a}-a \|_1,
\end{equation}
where $\alpha$ is a hyper parameter that balances dense code errors and sparse code errors. To optimize the weights in the observation layers and the reconstruction layers jointly, we train the whole model according to the overall loss (\ref{eq:overall_loss}) using gradient descent during backpropagation. 

Notice that using a standard toolbox, such as TensorFlow \cite{Tensorflow} would require both the observation and the reconstruction layers in Fig.~\ref{end-to-end} to be differentiable. This requirement brings us to the architecture shown in  Fig.~\ref{end-to-end-net}. The observation layers being a CNN is differentiable. For the reconstruction layer, we use the differentiable LISTA \cite{LISTA} architecture to compute approximate sparse codes. LISTA attempts to mimic the iterative shrinkage algorithm (ISTA) in a limited number of iterations ($T$) in a recurrent neural network architecture \cite{LISTA}. In the LISTA architecture shown in Fig.~\ref{end-to-end-net} has trainable parameters $W=D^T$ and $S=D^TD$. The detailed description of LISTA algorithm appears in \cite{LISTA}. Thus, the entire architecture is now differentiable and end-to-end trainable. We implemented this end-to-end trainable model using TensorFlow \cite{Tensorflow}.

We made End-to-End CNNCS more flexible and powerful by deriving a backpropagation rule for the reconstruction layer. Our backpropagation rule is \textit{independent} of the algorithm or architecture used for the forward pass in the reconstruction layer. For example, one can use a more powerful and accurate algorithm than LISTA or ISTA for the reconstruction layer. In fact, such an algorithm can also be \textit{non-differentiable} in nature. For the ease of implementation, we chose ISTA for all our experiments here. Our backpropagation rule assumes that the reconstruction layer solves the CS/SC optimization problem:
\begin{equation} \label{eq:sp}
\underset{\hat{a}}{\text{min}} \ \frac{1}{2}\| D\hat{a}-\hat{x} \|_2^{2}+\lambda\| \hat{a} \|_1 .
\end{equation}

Suppose $\delta \hat{a}$ and $\delta \hat{x}$ denote the partial derivatives of $L_1$ norm in the loss function (\ref{eq:overall_loss}) with respect to $\hat{a}$ and $\hat{x}$, respectively. Then the following backpropagation rule relates $\delta \hat{a}$ and $\delta \hat{x}$ (see Appendix for derivation):
\begin{equation} \label{eq:delx_body}
\delta \hat{x} =D \left(:, p \right)\left[D^{T}D\left( p,p\right)  \right] ^{-1}\delta \hat{a}\left(p \right),
\end{equation}
where $p = \left\{ i:\hat{a}_i\neq0 \right\}$ is the set of indices indicating non-zero components of $\hat{a}.$ $ D(:,p) $ indicates the columns of matrix $D$, whose indices belong to the set $p$. $D^{T}D\left( p,p\right)$ indicates the principal sub matrix of $D^{T}D$ with column and row indices belonging to set $p$. $\delta \hat{a}(p)$ denotes a vector comprising of only those elements of $\delta \hat{a}$ the indices for which belong to the set $p.$

In order to update the projection/sensing matrix $D,$ if $\delta D$ denotes the partial derivative of the $L_1$ norm term in the loss (\ref{eq:overall_loss}) with respect to $D,$ then, the following backpropagation rule can be derived (see Appendix):
\begin{equation} \label{eq:delD_body}
\begin{split}
\delta D \left(:, p \right)=\left(\hat{x}-D\hat{a} \right)\delta \hat{a}\left(p \right)^{T}\left[D^{T}D\left( p,p\right)  \right] ^{-1}- \\
D \left(:, p \right)\left[D^{T}D\left( p,p\right)  \right] ^{-1}\delta \hat{a}\left(p \right) \hat{a}\left(p \right)^{T},
\end{split}
\end{equation}
and
\begin{equation} \label{eq:delD_body1}
\delta D(:,q) = 0,
\end{equation}
where $q = \left\{ i:\hat{a}_i=0 \right\}$ is the set of indices indicating zero-elements of $\hat{a}.$
Based on (\ref{eq:delx_body}), (\ref{eq:delD_body}), (\ref{eq:delD_body1}), and (\ref{eq:overall_loss}) the system is able to backpropagate the error signal $(\hat{x}-x)+\alpha \delta \hat{x}$ to the CNN and optimize $D,$ so that the whole framework can be trained in an end-to-end fashion.

However, we note that the aforementioned rules (\ref{eq:delx_body}), (\ref{eq:delD_body}) may not be numerically stable and efficient for batch training mode, because it involves different matrices to be inverted for different images. Thus, we further derive  approximate, numerically stable, and computationally efficient backpropagation for batch training (see Appendix):
\begin{equation} \label{eq:delx_body2}
\delta \hat{x} \approx D \left(:, p \right) \delta \hat{a}\left(p \right),
\end{equation}
and
\begin{equation} \label{eq:delD_body2}
\delta D \left(:, p \right)\approx \left(\hat{x}-D\hat{a} \right)\delta \hat{a}\left(p \right)^{T} - D \left(:, p \right)\delta \hat{a}\left(p \right) \hat{a}\left(p \right)^{T}. 
\end{equation}

\section{Experiments}

\subsection{Details of Architecture}

We utilize a deep residual network (ResNet) \cite{ResNet} where we use its 152-layer model to predict the compressed cell location signal signal $\hat{x}$. This is referred as the ``observation layers'' in Fig.~\ref{end-to-end-net}. The loss function is defined as in (\ref{eq:overall_loss}). We perform fine-tuning on the weights in the fully-connected layer of the ResNet.

To further optimize our CNN model, we apply multi-task learning (MTL) \cite{Caruana1997}. Following MTL, in the $L_2$ norm of the loss function (\ref{eq:overall_loss}) two kinds of labels are provided. The first kind is the compressed cell vector: $x$, which carries the pixel location information for cell centroids in an image patch. The other kind is a scalar: cell count ($c$), which indicates the total number of cells in a training image patch. We concatenate the two kinds of labels into the final training label by $label = \left\lbrace x,\beta c\right\rbrace$, where $\beta$ is a hyper parameter. Then, $L_2$ loss is applied between the fusion $label$ and the output of CNN. Thus, supervision information for both cell detection and cell counting can be jointly used to fine-tune the parameters of ResNet.

Referring to Fig.~\ref{end-to-end-net}, we use LISTA \cite{LISTA} architecture for ``reconstruction layers'' in all our experiments. In the subsequent text, ``CNNCS'' refers to a model that uses a fixed Gaussian random projection matrix $D,$ which does not change during training. CNNCS model uses only the $L_2$ norm in the loss function (\ref{eq:overall_loss}). CNNCS has been used before in our previous work \cite{AAAI_workshop}, where a more complex algorithm DAL \cite{DAL} was used for reconstruction. By ``ECNNCS-1'' we refer to the end-to-end version of CNNCS that uses the full loss function (i.e., $L_2$ and $L_1$ norms) in (\ref{eq:overall_loss}). The backpropagation in ECNNCS-1 is based on the automatic differentiation by TensorFlow. Further, ``ECNNCS-2'' refers to the end-to-end CNNCS, where we use backpropagation rules (\ref{eq:delx_body2}) and (\ref{eq:delD_body2}).

\subsection{Datasets and Evaluation Criteria}


We utilize four cell datasets, on which CNNCS, ECNNCS-1, ECNNCS-2, faster RCNN \cite{faster-rcnn} and other comparison methods are evaluated. The 1st dataset is the ICPR 2012 mitosis detection contest dataset \cite{ICPR-2012} including 50 high-resolution (2084-by-2084) RGB microscope slides of Mitosis. The 2nd dataset \cite{ICPR-2014} is the ICPR 2014 grand contest of mitosis detection, which is a follow-up and an extension of the ICPR 2012 contest on detection of mitosis. Compared with the contest in 2012, the ICPR 2014 contest is more challenging because it contains many more images for training and testing. The 3rd dataset is the AMIDA-2013 mitosis detection dataset \cite{AMIDA-2013}, which contains 676 breast cancer histology images belonging to 23 patients. The 4th dataset is the AMIDA-2016 mitosis detection dataset \cite{AMIDA-2016}, which is an extension of the AMIDA 2013 contest on detection of mitosis. It contains 587 breast cancer histology images belonging to 73 patients for training, and 34 breast cancer histology images for testing with no ground truth available. For each dataset, the annotation that represents the location of cell centroids. Details of datasets are summarized in Table.~\ref{data-table}.

\begin{table}[h] \footnotesize
	\setlength{\abovecaptionskip}{0pt}
	\setlength{\belowcaptionskip}{0pt}
	\begin{center}
		\caption{\emph{Size} is the image size; \emph{Ntr/Nte} is the number of images selected for training and testing; \emph{AC} indicates the average number of cells per image.}
		\label{data-table}
		\begin{tabular}{*{22}{c}}
			\hline\noalign{\smallskip}
			Cell Dataset & Size & Ntr/Nte & AC\\
			\noalign{\smallskip}
			\hline
			\noalign{\smallskip}
			
			
			
			
			ICPR 2012 \cite{ICPR-2012} & 2084$\times$2084 & 35/15 & 5.31\\
			
			ICPR 2014 \cite{ICPR-2014} & 1539$\times$1376 & 1136/496 & 4.41\\
			
			AMIDA 2013 \cite{AMIDA-2013} & 2000$\times$2000 & 447/229 & 3.54\\
			
			AMIDA 2016 \cite{AMIDA-2016} & 2000$\times$2000 & 587/34 & 2.13\\
			
			\hline
		\end{tabular}
	\end{center}
\end{table}


For evaluation, we adopt the criteria of the ICPR 2012 mitosis detection contest \cite{ICPR-2012}, which is also adopted in several other cell detection contests. A detection would be counted as true positive ($TP$) if the distance between the predicted centroid and ground truth cell centroid is less than $\rho$. Otherwise, a detection is considered as false positives ($FP$). The missed ground truth cells are counted as false negatives ($FN$). In our experiments, $\rho$ is set to be the radius of the smallest cell in the dataset. Thus, only centroids that are detected to lie inside cells are considered correct. The results are reported in terms of Precision: $P=TP/(TP+FP)$ and Recall: $R=TP/(TP+FN)$ and $F_1$-score: $F_1=2PR/(P+R)$ in the following sections.

\subsection{Results}

\begin{figure}[h]
	\centering
	\setlength{\abovecaptionskip}{0pt}
	\setlength{\belowcaptionskip}{0pt}
 	\includegraphics[width=8.5cm]{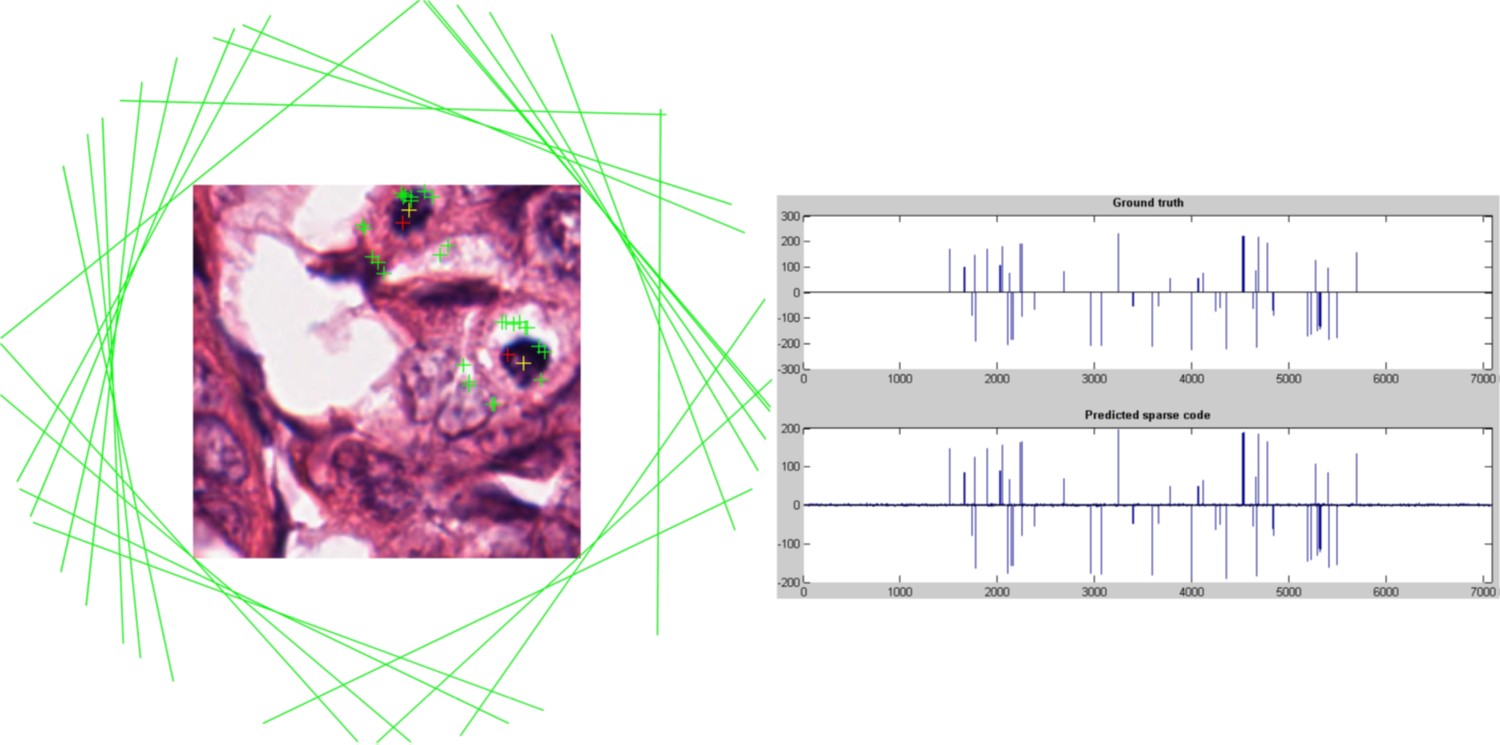}
 	\includegraphics[width=8.5cm]{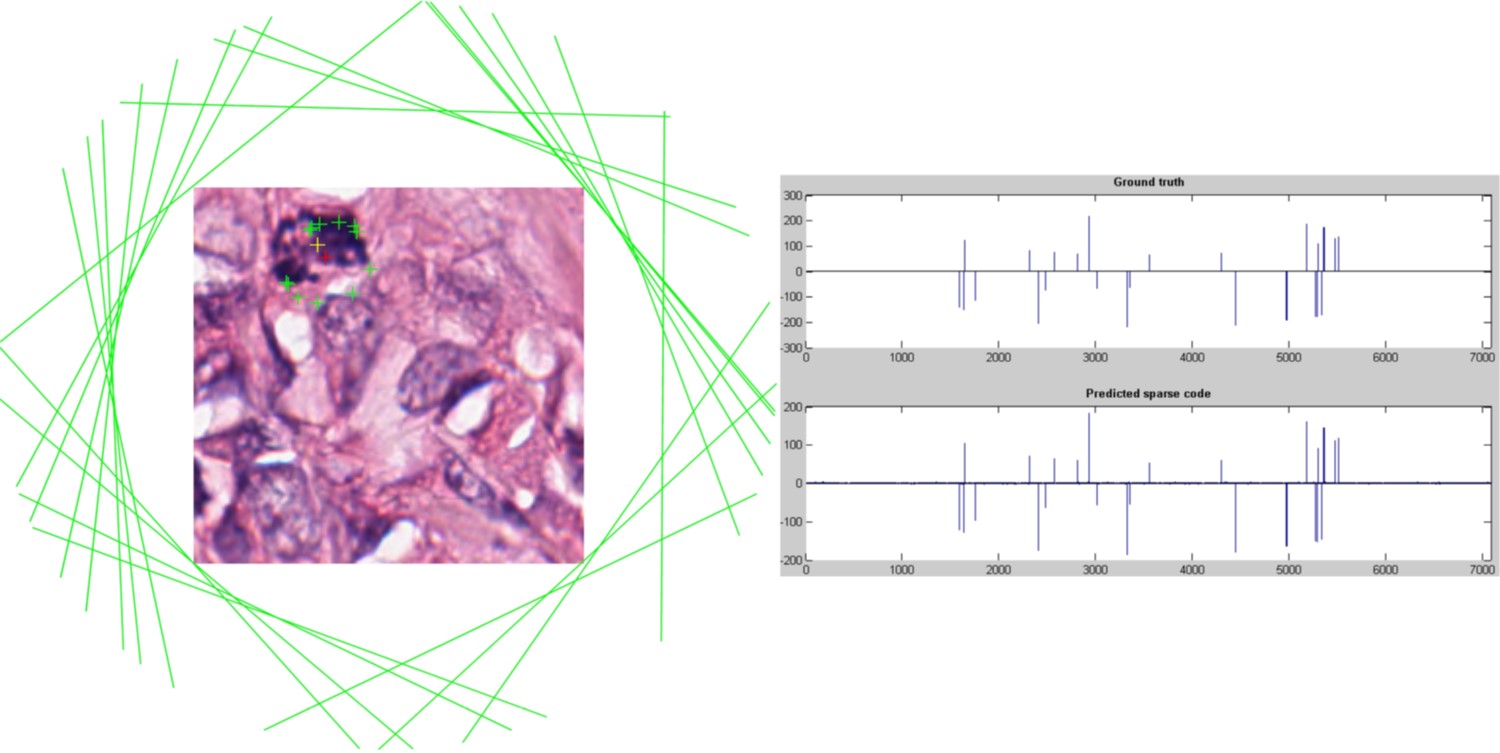}
	\caption{Example results on AMIDA-2016 dataset. Yellow cross indicates the ground-truth position of target cells. Green crosses indicate cell positions predicted by an observation axes. Red cross indicates the final detected cell position, which is the average of green crosses. On the right column, ground truth and predicted encoding for cell centers are shown.}
	\label{Example-results}
\end{figure}

In this section we describe experiments with the encoding obtained by projecting cell centroids onto randomly oriented straight lines as described in Section \ref{sec:encoding-scheme} (Fig. \ref{encoding1}). One of the advantages of this encoding is the redundancy introduced by several projections. This is shown in Fig. \ref{Example-results}. However, note that when more than one cells are located in a single patch, multiple predictions by multiple observations require clustering. Thus, we needed to perform mean shift clustering \cite{Meanshift} after recovery of the cell locations. The bandwidth for mean shift method is tuned as a hyper parameter. Mean shift method has the advantage of automatically estimating the number of clusters, given the bandwidth. After reconstructing signal $\hat{a},$ we apply a threshold value to it to prune out noises. Because the random lines are all located outside of the image patch, this threshold value has much less sensitivity to the accuracy, and we set it at 15 pixels.

We apply the proposed methods (CNNCS, ECNNCS-1 and ECNNCS-2) on the ICPR 2012 mitosis detection contest dataset, which consists of 35 training images and 15 testing images. For the training process, we extracted image sub-samples (260-by-260) with no overlap between each other from the 35 training images. After that three subsequent 90$^\circ$ rotations were performed on each sub-sample for data augmentation, resulting in a total of 8,960 training images. In addition, we perform random grid search to tune the three hyper parameters: $m=112, L=27, \beta=0.20, \lambda=0.39, \alpha=1.3,$ meanshift bandwidth=40. Finally, CNNCS gets the \textbf{highest} F1-score among all the comparison methods, details are summarized for the test set in Table~\ref{ICPR-2012-table}, where ECNNCS-2 showed best performance followed by a close second position for ECNNCS-1. Faster RCNN yielded considerable false positives in comparison that might be a result of imbalance in the training borne by a space discretization method. Our methods also handily defeated competition entries as shown in Table \ref{ICPR-2012-table}.
\begin{table}[h] \footnotesize
	\setlength{\abovecaptionskip}{0pt}
	\setlength{\belowcaptionskip}{0pt}
	\begin{center}
		\caption{Results of ICPR 2012 grand challenge of mitosis detection.}
		\label{ICPR-2012-table}
		\begin{tabular}{*{22}{c}}
			\hline\noalign{\smallskip}
			Method & Precision & Recall & F$_1$-score \\
			\noalign{\smallskip}
			\hline
			\noalign{\smallskip}
			
			UTRECHT & 0.511 & 0.680 & 0.584\\
			
			NEC \cite{NEC} & 0.747 & 0.590 & 0.659\\
			
			IPAL \cite{IPAL} & 0.698 & 0.740 & 0.718\\
			
			DNN \cite{DNN-IDSIA} &0.886&0.700&0.782\\
			
			RCasNN \cite{CasNN} &0.720&0.713&0.716\\
			
			CasNN-single \cite{CasNN} &0.738&0.753&0.745\\
			
			CasNN-average \cite{CasNN} &0.804&0.772&0.788\\
			
			\noalign{\smallskip}
			\cdashline{1-4}[11pt/3pt]
			\noalign{\smallskip}
			
			CNNCS &0.853 &0.791 &0.821 \\
			
			ECNNCS-1 &0.8637 &0.8385 &0.8509 \\
			
			ECNNCS-2 &0.8793 &0.8454 &0.8620\\
			
			Faster RCNN &0.5334 &0.8152 &0.6440\\
			
			\hline
		\end{tabular}
	\end{center}
\end{table}

In the next experiment, we use the ICPR 2014 contest of mitosis detection dataset (also called MITOS-ATYPIA-14) \cite{ICPR-2014}, which is a follow-up and an extension of the ICPR 2012 contest on detection of mitosis. Compared with the contest in 2012, the ICPR 2014 contest is much more challenging, which contains more images for training and testing. It provides 1632 breast cancer histology images, 1136 images for training, 496 images for testing. The test image labels have not been made public. Each image is 1539$\times$1376. We randomly divide the training images into training set (910 images) and validation set (226 images). We perform random search on the validation set to optimize the hyper parameters. The best performance on MITOS-ATYPIA-14 dataset is achieved when $m=103, L=30, \beta=0.24, \lambda=0.39, \alpha=1.5,$ meanshift bandwidth=40. On the test dataset, CNNCS \cite{AAAI_workshop} achieved the \textbf{highest} F$_1$-score among all the participated teams (see Table Table~\ref{ICPR-2014-table}). Because the test set is not available publicly and the competition is closed, we performed new experiments on the validation set with CNNCS (with LISTA architecture), ECNNCS-1, ECNNCS-2 and faster RCNN (see Table \ref{CNNCS-ISTA-ICPR-2014}). We note a similar trend for this dataset that ECNNCS-2 yielded the highest score and faster RCNN generated significant false positives. Also, Tables \ref{ICPR-2014-table} and \ref{CNNCS-ISTA-ICPR-2014} illustrate that the performance gap for CNNCS method on the validation and the test set is insignificant and thus, our models ECNNCS-1 and ECNNCS-2 may be expected to yield similar performances on the test set.

\begin{table}[h] \footnotesize
	\setlength{\abovecaptionskip}{0pt}
	\setlength{\belowcaptionskip}{0pt}
	\begin{center}
		\caption{Results of ICPR 2014 contest of mitosis detection in breast cancer histological images. F$_1$-scores of participated teams are shown.}
		\label{ICPR-2014-table}
		\begin{tabular}{*{22}{c}}
			\hline\noalign{\smallskip}
			Group & CUHK & MINES & YILDIZ & STRAS & CNNCS \cite{AAAI_workshop}\\
			\noalign{\smallskip}
			\hline
			\noalign{\smallskip}
			
			A06 &0.119&0.317&0.370&0.160&{\bf0.783}\\
			
			A08 &0.333&0.171&0.172&0.024&{\bf0.463}\\
			
			A09 &0.593&0.473&0.280&0.072&{\bf0.660}\\
			
			
			A19 &0.368&0.137&0.107&0.011&{\bf0.615}\\
			
			Average &0.356&0.235&0.167&0.024&{\bf0.633}\\
			
			\hline
		\end{tabular}
	\end{center}
\end{table}

\begin{table}[h] \footnotesize
	\setlength{\abovecaptionskip}{0pt}
	\setlength{\belowcaptionskip}{0pt}
	\begin{center}
		\caption{Performance methods on validation sets of ICPR-2014 dataset.}
		\label{CNNCS-ISTA-ICPR-2014}
		\begin{tabular}{*{22}{c}}
			\hline\noalign{\smallskip}
			Method & Precision & Recall & F1-score \\
			\noalign{\smallskip}
			\hline
			\noalign{\smallskip}
			
			CNNCS (validation set) &0.6210 &0.6527 &0.6365\\
			
			ECNNCS-1 (validation set) & 0.6421 &0.6745 &0.6579\\
			
			ECNNCS-2 (validation set) & 0.6384 &0.7056 &0.6703\\
			
			Faster RCNN (validation set) & 0.5351 & 0.7520 & 0.6253\\
			
			\hline
		\end{tabular}
	\end{center}
\end{table}

Our third experiment was performed on the AMIDA-2013 mitosis detection dataset, which contains 676 breast cancer histology images, belonging to 23 patients. Suspicious breast tissue is annotated by at least two expert pathologists, to label the center of each cancer cell. We train the proposed CNNCS method using 377 images, validate on 70 training images. The test set labels are not publicly available and the competition is closed. The image patch size was set to $200 \times 200$ to feed to the CNN. The best performance on the validation set of AMIDA-2013 dataset is achieved when $m=118, L=25, \beta=0.32, \lambda=0.43, \alpha=1.4,$ meanshift bandwidth=45. Table \ref{AMIDA-2013-table} shows these results.

\begin{table}[h] \footnotesize
	\setlength{\abovecaptionskip}{0pt}
	\setlength{\belowcaptionskip}{0pt}
	\begin{center}
		\caption{Top 5 Results of AMIDA-2013 MICCAI grand challenge of mitosis detection. Our experiments appear below the dotted line.}
		\label{AMIDA-2013-table}
		\begin{tabular}{*{22}{c}}
			\hline\noalign{\smallskip}
			Method & Precision & Recall & F$_1$-score \\
			\noalign{\smallskip}
			\hline
			\noalign{\smallskip}
			
			IDSIA \cite{DNN-IDSIA} &0.610&0.612&0.611\\
			
			DTU &0.427&0.555&0.483\\
			
			AggNet \cite{AggNet16} &0.441&0.424&0.433\\
			
			CUHK &0.690&0.310&0.427\\
			
			SURREY &0.357&0.332&0.344\\
			
			
			
			
			
			
			
			
			
			
			
			\noalign{\smallskip}
			\cdashline{1-4}[11pt/3pt]
			\noalign{\smallskip}
						
			CNNCS (on validation set) &0.3952&0.5879&0.4727\\
		
			ECNNCS-1 (on validation set) &0.5988 &0.6028 &0.6008 \\
		
			ECNNCS-2 (on validation set) &0.6137 &0.6541 &0.6332\\
		
			Faster RCNN (on validation set) &0.4875  &0.7319 &0.5852\\
			\hline
		\end{tabular}
	\end{center}
\end{table}

Our fourth experiment is on the AMIDA-2016 mitosis detection challenge (also called TUPAC16), which is a follow-up and an extension of the AMIDA-2013 contest on detection of mitosis. This competition has been closed and test image labels are not available publicly. Table \ref{AMIDA-2016-table} illustrates results. Same set of hyper parameters as in AMIDA-2013 were used here. In addition to the F$_1$ scores, Table \ref{CNNCS-ISTA-AMIDA-2016} shows precision and recall numbers for proposed methods and faster RCNN. Once again, we notice that faster RCNN generated lots of false positives bringing down the F$_1$-score significantly. Fig.~\ref{Example-results} shows two example results on AMIDA-2016 dataset.

\begin{table}[h] \footnotesize
	\setlength{\abovecaptionskip}{0pt}
	\setlength{\belowcaptionskip}{0pt}
	\begin{center}
		\caption{Top 5 Results of AMIDA-2016 MICCAI grand challenge of mitosis detection. Our experiments appear below the dotted line.}
		\label{AMIDA-2016-table}
		\begin{tabular}{*{22}{c}}
			\hline\noalign{\smallskip}
			Team & F$_1$-score \\
			\noalign{\smallskip}
			\hline
			\noalign{\smallskip}
			
			Lunit Inc. &0.652\\
			
			IBM Research Zurich and Brazil &0.648\\
			
			Contextvision (SLDESUTO-BOX) &0.616\\
			
			The Chinese University of Hong Kong &0.601\\
			
			Microsoft Research Asia &0.596\\
			
			
			
			
			
			
			
			
			\noalign{\smallskip}
			\cdashline{1-4}[11pt/3pt]
			\noalign{\smallskip}
			
			CNNCS (on validation set) &0.6264\\
            
			ECNNCS-1 (on validation set) &0.6576\\
		
			ECNNCS-2 (on validation set) &0.6856\\
            
			Faster RCNN (on validation set) &0.6043\\
			\hline
		\end{tabular}
	\end{center}
\end{table}

\begin{table}[htbp] \footnotesize
	\setlength{\abovecaptionskip}{0pt}
	\setlength{\belowcaptionskip}{0pt}
	\begin{center}
		\caption{Performance of end-to-end trainable models on AMIDA-2016 dataset.}
		\label{CNNCS-ISTA-AMIDA-2016}
		\begin{tabular}{*{22}{c}}
			\hline\noalign{\smallskip}
			Method & Precision & Recall & F1-score \\
			\noalign{\smallskip}
			\hline
			\noalign{\smallskip}
			
			CNNCS (on validation set)  &0.5478 &0.7314 &0.6264\\
			
			ECNNCS-1 (on validation set)  & 0.5936 &0.7370 &0.6576\\
			
			ECNNCS-2 (on validation set)  & 0.6423 &0.7352 &0.6856\\
			
			Faster RCNN (on validation set) & 0.4961 & 0.7728 & 0.6043\\
			
			\hline
		\end{tabular}
	\end{center}
\end{table}


\section{Conclusions}

We have presented an end-to-end training method for cell detection from microscopy images using convolutional neural network and compressed sensing (sparse coding). Experiments hold our premises that (1) end-to-end training paradigm is more powerful than a non-end-to-end counterpart, and (2) compression is more promising than space discretization, which all methods to date use for cell detection. This is the first proposed method that derives a practical and algorithm independent backpropagation rule for compressed sensing/sparse coding. This backpropagation rule can be used beyond the presented application here. 



\appendices
\section{Backpropagation for reconstruction layer}


A reconstruction layer for CS/SP finds a vector $a$ according to the following minimization:
\begin{equation} \label{eq:l1}
\arg\min_{a}\frac{1}{2}\parallel x- Da\parallel_{2}^{2} +\lambda\parallel a\parallel_{1},
\end{equation}
where $D\in \mathbb{R}^{m\times n}$, $a\in \mathbb{R}^{n\times 1}$, $x\in \mathbb{R}^{m\times 1}$. Thus, according to (\ref{eq:l1}), $x$ and $a$ are input and output for the reconstruction layer, respectively. $D$ is a parameter in the this layer.

For the minimization problem (\ref{eq:l1}), there is no known closed form solution. Thus, to derive backpropagation for the reconstruction layer, we replace the $L_1$ norm by a smooth, convex approximation $f\left(a \right) $. An example smooth approximation is $f(a) = \sum_{i=1}^{n} \sqrt[]{a_i^2+\epsilon^2}$, where $\epsilon$ is a small real number. So, the minimization problem (\ref{eq:l1}) becomes:
\begin{equation} \label{eq:l1_approx}
\arg\min_{a}\frac{1}{2}\parallel x- Da \parallel_{2}^{2} +\lambda f\left(a \right). 
\end{equation} 
For an $a$ that minimizes (\ref{eq:l1_approx}), the necessary and sufficient condition is given by:
\begin{equation} \label{eq:nsc}
D^{T}\left(Da-x \right)+\lambda \nabla f\left(a \right) = 0.
\end{equation}

\subsection{Derivation of Partial Derivative $\delta x$}

Differentiating equation (\ref{eq:nsc}) with respect to $x$, we get
\begin{equation} \label{eq:jaca}
\left[J_{x}^{a}\right]^{T}= D\left[D^{T}D+\lambda H  \right]^{-1},
\end{equation}
where $H$ is the Hessian of $f$ and $J_x^a$ is the Jacobian. So, we have:
\begin{equation} \label{eq:delx}
\delta x = \left[J_{x}^{a}\right]^{T} \delta a = D\left[D^{T}D+\lambda H  \right]^{-1} \delta a,
\end{equation}
where $\delta x$ and $\delta a$ are partial derivatives of the neural network loss function with respect to $x$ and $a,$ respectively.

Let $p=\left\lbrace i:a_{i}\neq0 \right\rbrace $ and $q=\left\lbrace i:a_{i}=0 \right\rbrace $. Further, let $A\left(:,p\right)$ denote the columns of $A$, whose indices belong to the set $p$. Similarly, $A(q,p)$ denotes the sub-matrix of $A$, where row and column indices belong to the sets $q$ and $p$, respectively. With these notations, we rewrite (\ref{eq:jaca}) as:
\begin{equation} \label{eq:jaca1}
\begin{split}
& \begin{bmatrix}
J_{x}^{a}\left(p,: \right) 
 \\ J_{x}^{a}\left(q,: \right) 
\end{bmatrix}^T= \begin{bmatrix}
D\left(:,p \right) 
& D\left(:,q \right) 
\end{bmatrix}
\\
& \begin{bmatrix}
D^{T}D\left(p,p \right)+\lambda H\left(p,p \right)  &  D^{T}D\left(p,q\right) \\
D^{T}D\left(q,p\right) & D^{T}D\left(q,q \right)+\lambda H\left(q,q \right)
\end{bmatrix}^{-1}\\
&= \\
&\begin{bmatrix}
I_{p}   &  0 \\
-\left[ D^{T}D\left(q,q \right)+\lambda H\left(q,q \right)\right]^{-1}D^{T}D\left(q,p\right) & I_{q}
\end{bmatrix}\\
&\begin{bmatrix}
U^{-1}    & 0  \\
0 & \left[ D^{T}D\left(q,q \right)+\lambda H\left(q,q \right)\right]^{-1}
\end{bmatrix} \\
&\begin{bmatrix}
I_{p} & -D^{T}D(p,q)\left[ D^{T}D\left(q,q \right)+\lambda H\left(q,q \right)\right]^{-1}\\
0 &I_{q}
\end{bmatrix},
\end{split}
\end{equation}
where, we used Schur complement in (\ref{eq:jaca1}) and noted that $H$ is a diagonal matrix. $I_{p}$ and $I_{q}$ are identity matrices of order $\lvert p\lvert$ and $\lvert q\lvert$, respectively. $U$ is defined as:
\begin{flalign}
\begin{split}
& U=D^{T}D\left(p,p \right)+\lambda H\left(p,p \right)- \\
& D^{T}D\left(p,q\right)\left[ D^{T}D\left(q,q \right)+ \lambda H\left(q,q \right)\right]^{-1}D^{T}D\left(q,p\right).
\end{split}
\end{flalign}
Note that $\lambda H\left(p,p \right)\rightarrow 0.$ Further,
$\lambda H\left(q,q \right)$ is a diagonal matrix with diagonal entries tending to infinity. This property is common for any reasonable $L_1$ norm approximation. For example, the second derivative of $f_{i}\left(a_i \right)=\sqrt{a_i^{2}+\epsilon^2} $ is $f^{''}_{i}\left(a_i\right)=\dfrac{1}{\sqrt{a_i^{2}+\epsilon^2}}-\dfrac{a_i^{2}}{\left( a_i^{2}+\epsilon^2\right)^{3/2}} $. Thus, the second derivative satisfies:
$f^{''}_i\left(a_i \right) \rightarrow \infty$, when $\epsilon \to 0,$ and $a_i=0$; and $f^{''}_i\left(a_i \right) \rightarrow 0$ when $\epsilon \to 0,$ and $a_i\neq 0$. Therefore, we also have:


\begin{equation}
\left[ D^{T}D\left(q,q \right)+\lambda H\left(q,q \right)\right] ^{-1}\rightarrow 0.
\end{equation}
Combining these results, we get:
\begin{equation}
\begin{split}
& \begin{bmatrix}
D^{T}D\left(p,p \right)+\lambda H\left(p,p \right)   &  D^{T}D\left(p,q\right) \\
D^{T}D\left(q,p\right) & D^{T}D\left(q,q \right)+\lambda H\left(q,q \right)
\end{bmatrix}^{-1}\rightarrow \\
& \begin{bmatrix}
\left[ D^{T}D\left(p,p \right)\right] ^{-1}   &  0 \\
0 & 0
\end{bmatrix}.
\end{split}
\end{equation}
Note that for a random Gaussian matrix $D,$ $D^TD(p,p)=D^T(p,:)D(:,p)$ is called a Wishart matrix and assuming $m \geq |p|,$ it is invertible \cite{Eaton2007}. Using these results in (\ref{eq:jaca1}), we get:
\begin{flalign} \label{eq:jaca2}
&\begin{bmatrix}
J_x^a\left(p,: \right) \\ J_x^a\left(q,: \right) 
\end{bmatrix}^T = \nonumber \\
&\begin{bmatrix}
D\left(:,p \right) 
& D\left(:,q \right) 
\end{bmatrix} \begin{bmatrix}
\left[ D^{T}D\left(p,p \right)\right] ^{-1}   &  0 \\
0 & 0
\end{bmatrix} = \nonumber\\
&\begin{bmatrix}
D\left(:,p \right)\left[ D^{T}D\left(p,p \right)\right] ^{-1} 
&0
\end{bmatrix}. 
\end{flalign}
Using (\ref{eq:jaca2}) in (\ref{eq:delx}), we finally obtain: 
\begin{equation} \label{eq:finalx}
\delta x=
D\left(:,p \right)\left[ D^{T}D\left(p,p \right)\right] ^{-1}\delta a\left(p \right),
\end{equation}
where $\delta a(p)$ denotes a vector comprising of only those elements of $\delta a,$ the indices for which belong to set $p.$

\subsection{Derivation of Partial Derivative $\delta D$}
Let us partition the $m\times n$ projection matrix as: $D_{m\times n}=\begin{bmatrix}
D_{1}& D_{2}& \ldots & D_{n} \end{bmatrix}$, where $D_i$ is a $m\times 1$ column vector. With these notations, equation (\ref{eq:nsc}) can be written as:
\begin{equation} \label{eq:nscd}
D_{i}^{T}\sum_{j=1}^{n}D_{j}a_{j}+\lambda f^{'}\left( a_{i}\right) =D_{i}^{T}x,\ \ i=1,2,\ldots, n,
\end{equation}
where $f^{'}\left( a_{i}\right)$ denotes the $i^{th}$ component of $\nabla f.$
Differentiating (\ref{eq:nscd}) with respect to $D_i$, we obtain:
\begin{equation} \label{eq:eqi}
\begin{split}
&\sum_{j=1}^{n}D_{j}a_{j}+D_{i}a_{i}+\sum_{j=1}^{n}D_{i}^{T}D_{j}\dfrac{\partial a_{j}}{\partial D_{i}}+\\
& \lambda f^{''}\left( a_{i}\right)\dfrac{\partial a_{i}}{\partial D_{i}} =x, \ \  i=1,2,\ldots, n,
\end{split}
\end{equation}
where $f^{''}\left( a_{i}\right)$ denotes the $i^{th}$ diagonal entry of the Hessian $H.$
Differentiating (\ref{eq:nscd}) with respect to $D_k$, where $k\neq i,$ we obtain:
\begin{flalign} \label{eq:Da}
\begin{split}
&D_{i}a_{k}+\sum_{j=1}^{n}D_{i}^{T}D_{j}\dfrac{\partial a_{j}}{\partial D_{k}}+\\
&\lambda f^{''}\left( a_{i}\right)\dfrac{\partial a_{i}}{\partial D_{k}} =0,\left\{
\begin{array}{ll}
i,k=1,2,\ldots, n,\\
i\neq k.
\end{array} \right.
\end{split}
\end{flalign}
Interchanging indices $i$ and $k$ in (\ref{eq:Da}), we obtain:
\begin{flalign} \label{eq:eqk}
\begin{split}
&D_{k}a_{i}+\sum_{j=1}^{n}D_{k}^{T}D_{j}\dfrac{\partial a_{j}}{\partial D_{i}}+\\
&\lambda f^{''}\left( a_{k}\right)\dfrac{\partial a_{k}}{\partial D_{i}} =0,\left\{
\begin{array}{ll}
i,k=1,2,\ldots, n,\\
i\neq k.
\end{array} \right.
\end{split}
\end{flalign}
Combining (\ref{eq:eqi}) and (\ref{eq:eqk}), we obtain:
\begin{equation}
\begin{split}
& \dfrac{\partial a}{\partial D_i}\left[D^{T}D+\lambda H \right] = \\
& -\begin{bmatrix} D_{1}a_{i}&D_{2}a_{i}&\ldots&Da+D_{i}a_{i}-x&\ldots&D_{n}a_{i}\end{bmatrix}, \\
& i=1,2,\ldots, n.
\end{split}
\end{equation}
Here, $\dfrac{\partial a}{\partial D_{i}}$ is defined as:
\begin{equation}
\left( \dfrac{\partial a}{\partial D_{i}}\right) _{m\times n} =\begin{bmatrix} \dfrac{\partial a_{1}}{\partial D_{i}}&\dfrac{\partial a_{2}}{\partial D_{i}}&\ldots&\dfrac{\partial a_{n}}{\partial D_{i}}\end{bmatrix} _{m\times n}
\end{equation}
So, we obtain:
\begin{equation}
\begin{split}
\dfrac{\partial a}{\partial D_{i}}= & -\begin{bmatrix} D_{1}a_{i}&D_{2}a_{i}&\ldots&Da+D_{i}a_{i}-x&\ldots&D_{n}a_{i}\end{bmatrix} \\
& \left[ D^{T}D+\lambda H\right] ^{-1}, \ \ i=1,2,\ldots, n.
\end{split}
\end{equation}
Now using the chain rule, we get:
\begin{equation} \label{eq:neqn}
\begin{split}
&\delta D_{i} \equiv \bigtriangledown_{Di}\left(network \ loss \right) =\left(\dfrac{\partial a}{\partial D_{i}} \right)_{m\times n}\delta a_{n\times 1}= \\
&-\begin{bmatrix} D_{1}a_{i}&D_{2}a_{i}&\ldots Da+D_{i}a_{i}-x&\ldots&D_{n}a_{i}\end{bmatrix} \\
& \left[ D^{T}D+\lambda H\right] ^{-1}\delta a, \ \ i=1,2,\ldots, n.
\end{split}
\end{equation}

Collecting these $n$ vector equations (\ref{eq:neqn}) in a matrix, we get:
\begin{equation} \label{eq:delD0}
\begin{split}
\delta D=\left(x-Da \right) \delta a^{T}\left[ D^{T}D+\lambda H\right] ^{-1} -\\
D\left[ D^{T}D+\lambda H\right] ^{-1}\delta aa^{T}.
\end{split}
\end{equation}
Using the previous results of Schur complement in (\ref{eq:delD0}), we finally arrive at:
\begin{equation} \label{eq:finalD1}
\begin{split}
\delta D(:,p) =\left(x-Da \right) \delta a\left(p \right) ^{T}\left[D^{T}D\left(p,p \right)  \right]^{-1}- \\
D(:,p)\left[D^{T}D\left(p,p \right)  \right]^{-1} \delta a\left(p \right)a(p)^{T},
\end{split}
\end{equation}
and 
\begin{equation} \label{eq:finalD2}
\delta D(:,q) = 0.
\end{equation}

\subsection{Efficient Backpropagation for Batch Training}

The backpropagation rules (\ref{eq:finalx}) and (\ref{eq:finalD1}) are not computationally efficient for batch training, where a number of images are fed to the system at a time. The source of the computational inefficiency is the dependency of the matrix inversion $[D^TD(p,p)]^{-1}$ on the set of indices $p$, which is different for different sparse codes $a.$ This observation led us to develop an approximate, batch friendly gradient computation.

For a random matrix $D_{m \times n}$ with entries drawn independently from a zero mean Gaussian with variance $1/m$, the matrix $[D^TD(p,p)]^{-1}$ follows an inverse Wishart distribution with mean proportional to an identity matrix \cite{Eaton2007}. Thus, using a first order approximation (i.e., representing a random variable with its mean), we replace $[D^TD(p,p)]^{-1}$ with an identity matrix in (\ref{eq:finalx}) and (\ref{eq:finalD1}) and obtain approximate, numerically stable, computationally efficient and batch friendly backpropagation rules:
\begin{equation}
\delta x \approx D \left(:, p \right) \delta a\left(p \right),
\end{equation}
and
\begin{equation}
\delta D \left(:, p \right) \approx \left(x-Da \right)\delta a\left(p \right)^{T} - D \left(:, p \right)\delta a\left(p \right) a\left(p \right)^{T}. 
\end{equation}

\bibliographystyle{IEEEtran}
\bibliography{refs}



%








\end{document}